\documentclass[iicol, sn-basic, Numbered]{sn-jnl}

\usepackage{graphicx}%
\usepackage{multirow}%
\usepackage{amsmath,amssymb,amsfonts}%
\usepackage{amsthm}%
\usepackage{mathrsfs}%
\usepackage[title]{appendix}%
\usepackage{xcolor}%
\usepackage{textcomp}%
\usepackage{manyfoot}%
\usepackage{booktabs}%
\usepackage{algorithm}%
\usepackage{algorithmicx}%
\usepackage{algpseudocode}%
\usepackage{listings}%
\usepackage{mathtools}
\usepackage{lipsum}
\usepackage{algorithm}
\usepackage{algpseudocode}
\usepackage{caption}
\usepackage{url} 

\usepackage{enumitem}
\usepackage{siunitx}
\usepackage{kotex}
\usepackage{booktabs} 
\usepackage{color}
\usepackage{wrapfig}
\usepackage[most]{tcolorbox}
 \usepackage{ulem}

\theoremstyle{thmstyleone}%
\theoremstyle{thmstyletwo}%

\theoremstyle{thmstylethree}%

\newcommand{\method}{{\textsc{C2F-Space}}}

\newtcolorbox{promptblock}[1][]{%
  breakable,
  colback=gray!5,
  colframe=black,
  sharp corners,
  boxrule=0.5pt,
  width=0.97\linewidth,
  before=\vspace{1pt},
  after=\vspace{1pt},
  before upper={\fontsize{9}{11}\selectfont},
  #1
}

\begin{document}

\title[C2F-Space]{
\begin{center}
\textsc{C2F-Space}: Coarse-to-Fine Space Grounding\\ 
for Spatial Instructions using Vision-Language Models
\end{center}
}

\author[1]{\fnm{Nayoung} \sur{Oh}}\email{nyoh@kaist.ac.kr}

\author[1]{\fnm{Dohyun} \sur{Kim}}\email{dohyun141@kaist.ac.kr}

\author[1]{\fnm{Junhyeong} \sur{Bang}}\email{jazzdosa@kaist.ac.kr}

\author*[2]{\fnm{Rohan} \sur{Paul}}\email{rohan@cse.iitd.ac.in}

\author*[1]{\fnm{Daehyung} \sur{Park}}\email{daehyung@kaist.ac.kr}

\affil[1]{\orgname{Korea Advanced Institute of Science and Technology}, \orgaddress{\state{Daejeon}, \country{Republic of Korea}}}

\affil[2]{\orgname{Indian Institute of Technology Delhi}, \orgaddress{\state{Hauz Khas}, \country{India}}}


\abstract{
Space grounding refers to localizing a set of spatial references described in natural language instructions. Traditional methods often fail to account for complex reasoning\textemdash such as distance, geometry, and inter-object relationships\textemdash while vision-language models (VLMs), despite strong reasoning abilities, struggle to produce a fine-grained region of outputs. To overcome these limitations, we propose \method, a novel coarse-to-fine space-grounding framework that (i) estimates an approximated yet spatially consistent region using a VLM, then (ii) refines the region to align with the local environment through superpixelization. For the coarse estimation, we design a grid-based visual-grounding prompt with a \textit{propose-validate} strategy, maximizing VLM's spatial understanding and yielding physically and semantically valid canonical region (i.e., ellipses). For the refinement, we locally adapt the region to surrounding environment without over-relaxed to free space. 
We construct a new space-grounding benchmark and compare \method\ with five state-of-the-art baselines using success rate and intersection-over-union. Our \method\ significantly outperforms all baselines. Our ablation study confirms the effectiveness of each module in the two-step process and their synergistic effect of the combined framework. We finally demonstrate the applicability of \method\ to simulated robotic pick-and-place tasks.
}
\keywords{Space Grounding, Vision-Language Model, Coarse-to-fine, Spatial Reasoning}

\maketitle

\section{Introduction}
\begin{figure*}[t]
    \centering
    \includegraphics[width=\textwidth]{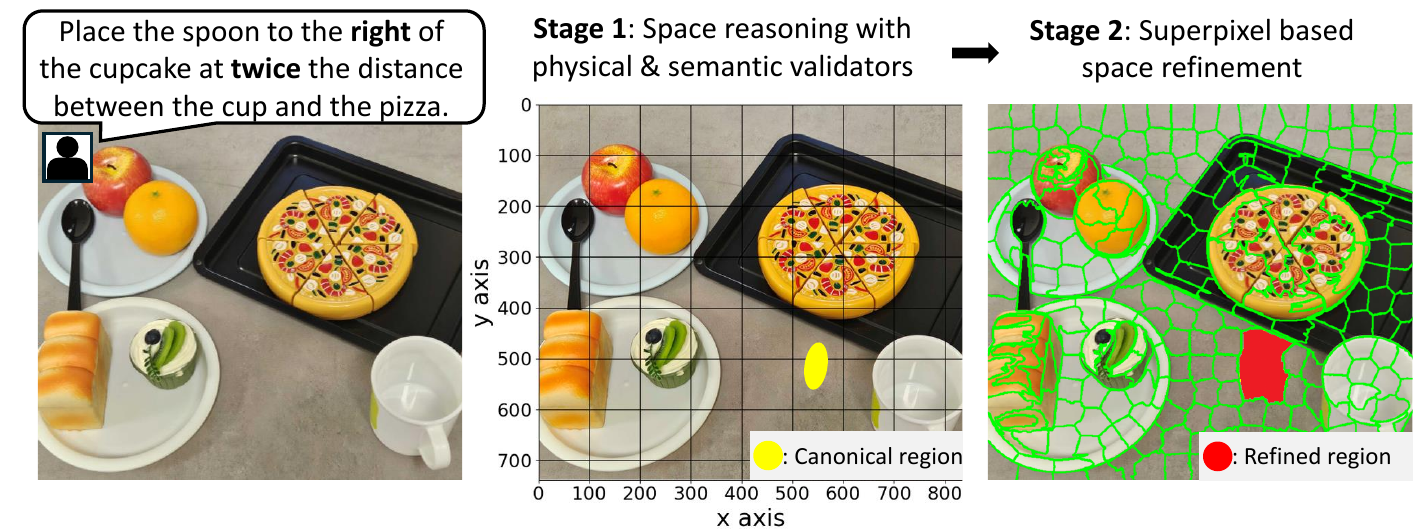}
    \caption{Illustration of the two-stage space grounding result produced by the proposed \method . The grid-guided prompt enables the VLM to generate a coarse region proposal (e.g., an ellipsoid) through spatially multiplicative reasoning. A superpixel-based enhancement process then refines this proposal into a fine-grained spatial mask.}
    \label{fig:enter-label}
\end{figure*}
Space grounding refers to the process of mapping linguistic expressions to spatial regions within an environment~\cite{kim2024lingo}. The process often requires complex spatial reasoning that accounts for distance, geometry, and inter-object relationships, which have yet to be thoroughly investigated. Fig.~\ref{fig:enter-label} illustrates a representative example in a robotic pick-and-place scenario, where a human provides an instruction: ``Place the spoon to the right of the cupcake at twice the distance between the cup and the pizza.'' The interpretation of this instruction requires not only estimating the distance, but also reasoning about proportional relationships to determine the target position among candidates located twice that distance to the right of the cupcake.

Early approaches link simple spatial expressions (e.g., `near') to a limited category of segments (e.g., `next to the stop')~\cite{jain2023ground}. Subsequent approaches support compositional expressions~\cite{zhao2023differentiable,gkanatsios2023energybased}. Recently, \citet{kim2024lingo} introduce a probabilistic update mechanism to resolve the ambiguity in compositional expressions. Despite these advances, their generalization remains limited due to the small scale of annotated datasets. 

With advances in large language models (LLMs) and vision-language models (VLMs), researchers begin leveraging large pre-trained models to enhance spatial reasoning. Notable examples include \textsc{RoboPoint}~\cite{yuan2025robopoint}, which fine-tunes a VLM to localize target regions as coarse point sets, and recent VLMs, such as Molmo~\cite{deitke2025molmo} and Gemini 2.5~\cite{comanici2025gemini}, demonstrate zero-shot 2D point grounding~\cite{cheng2025pointarena}. Nevertheless, the coarse, point-level nature of these outputs often lacks the rigorous spatial precision required for downstream applications, particularly in fine-grained robotic manipulation. Although visual prompting methods provide more detailed guidance and complement text instructions~\cite{shtedritski2023does, cai2024vip}, they typically rely on coarse visual markers (e.g., circles), which limit their fine-grained spatial reasoning capability.

Therefore, we propose \method, a novel coarse-to-fine space-grounding method that infers semantically and geometrically valid space given spatial instructions. Our method consists of two stages: the first stage enables the VLM to reason about approximate but spatially complex relationships \textemdash such as geometric relations, relational distances, and reference uniqueness\textemdash through grid-based inpainting image prompting. This then proposes a set of ellipsoidal region candidates, validating them to ensure both physical feasibility for placement actions and semantic consistency with sub-components of the spatial instruction. In the second stage, \method\ locally adapts the coarse candidates to precisely align with the surrounding environment using superpixelization, while avoiding over-relaxed regions that extend into free space.

We evaluate \method\ on a newly constructed space-grounding benchmark consisting of $350$ problems. The benchmark includes spatial instructions featuring diverse relational distances and reference ambiguities, such as single- and multi-hop relations as well as unique and non-unique references\textemdash for example, ``below the red bottle by half the distance between the basket and the banana." Our proposed \method\ outperforms state-of-the-art spatial grounding baselines~\cite{shridhar2022cliport, kim2024lingo, deitke2025molmo, yuan2025robopoint} and vision-language models~\cite{deitke2025molmo, openai2025o4}, complementing the grounding process of o4-mini~\cite{openai2025o4}. Finally, we demonstrate the applicability of \method\ in simulated robotic pick-and-place tasks within PyBullet~\cite{pybullet} environments.

Our key contributions are as follows:
\begin{itemize}[leftmargin=*, noitemsep,topsep=0pt]
    \item We introduce a grid-guided space-reasoning prompt, which allows a VLM to propose and validate physically feasible and semantically consistent region candidates.
    \item We provide a superpixel-based refinement module that adjusts regional candidates for fine-grained alignment with the surrounding environment, while preventing overextension into free space.
    \item We construct a new space-grounding benchmark of $350$ examples consisting of diverse challenging instructions performing extensive comparisons with state-of-the-art spatial grounding baselines.
\end{itemize}

\section{Related Work}

\textbf{Spatial Understanding}: Spatial understanding is a fundamental capability for intelligent machines, enabling downstream decision-making and execution in real-world environments. Early approaches employ rule-based mapping, which directly associates spatial expressions with sensory patterns to recognize environmental layouts~\cite{skubic2004spatial, zender2008conceptual}. Studies then adopt discriminative probabilistic models, such as support vector machines, to map spatial description with visual observations~\cite{lan2012image, kulkarni2013babytalk}. However, these methods capture only predefined spatial relations and fail to represent the diversity necessary for comprehensive scene understanding ~\cite{lu2016visual}.

Researchers then adopt large language models (LLMs) in combination with visual inputs, forming vision-language models (VLMs) that learn spatial semantics from diverse, large-scale multimodal datasets~\cite{lu2019vilbert,radford2021learning,tan2019lxmert}. 
VLMs provide open-world multimodal understanding, making them applicable to a broad range of downstream tasks, such as image-text retrieval~\cite{chen2022pali}, zero-shot visual question answering~\cite{li2023blip}, and segmentation~\cite{lai2024lisa}. However, early VLMs fail spatial reasoning since they behave as a bag-of-tokens, which lose positional detail~\cite{yuksekgonul2023and,li2024topviewrs,chen2024spatialvlm}. To overcome the limitation, recent approaches integrate depth features into VLMs to provide scale cues~\cite{cheng2024spatialrgpt}. 
Furthermore, fine-tuning VLMs using extensive spatial relation annotations improves reasoning over complex object interactions in diverse scenes~\cite{yuan2025robopoint, song2025robospatial}. Alternatively, without direct modification of existing VLMs, our method guides a VLM with structured visual and textual prompts, enhancing its ability to understand spatial expressions and predict the target space described by the instruction.

\noindent\textbf{Spatial grounding}: Spatial grounding methods map linguistic spatial expressions to physical points or areas introducing various representations. Early approaches manually associate predefined predicates with structured representations, such as potential fields~\cite{stopp1994utilizing} or fuzzy spatial membership functions~\cite{bloch2003representation,tan2014grounding}. To overcome the limitations of these hand-crafted mappings, neural network-based methods predict either pixel coordinates for target placement~\cite{venkatesh2021spatial} or dense probability maps over the image~\cite{mees2020learning, shridhar2022cliport}. To capture more complex distributions, recent studies adopt parameterized probability representations, such as Gaussian mixture models~\cite{zhao2023differentiable} and Boltzmann energy functions~\cite{gkanatsios2023energybased}. Extending to spatiotemporal compositional reasoning, LINGO-Space models the target space via a Bayesian update of polar distributions~\cite{kim2024lingo}.

\begin{figure*}[t]
    \centering
    \includegraphics[width=\linewidth]{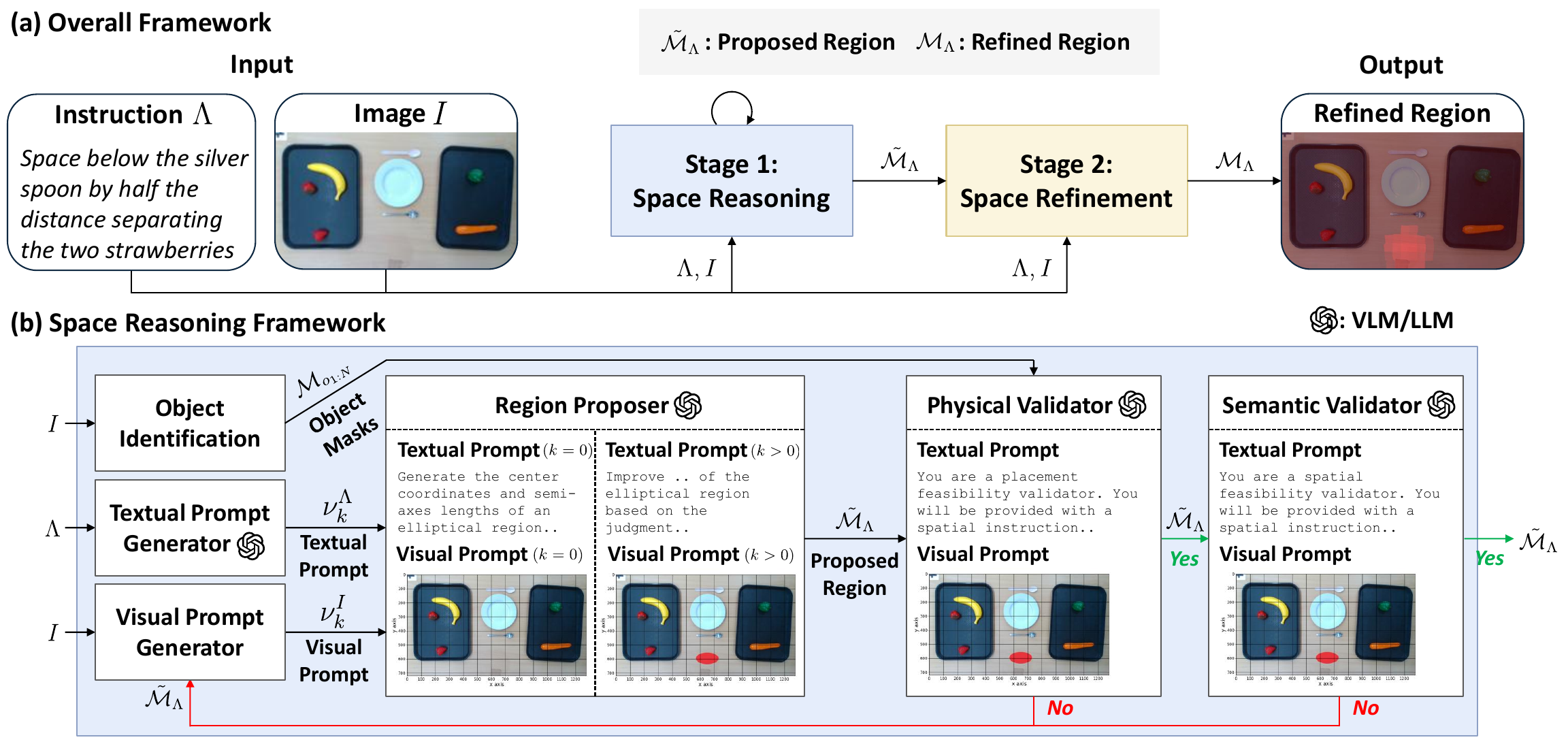}
    \caption{
    (a) Overall framework of \method. Given an instruction $\Lambda$ and an image $I$, the space‑reasoning module proposes a candidate region $\tilde{\mathcal{M}}_{\Lambda}$, and the subsequent space‑refinement module adapts this proposal to produce the precise region $\mathcal{M}_{\Lambda}$.
    (b) The space-reasoning module iteratively proposes and validates a canonical spatial region $\tilde{\mathcal{M}}_{\Lambda}$. At each iteration $k$, the VLM-based proper predicts an elliptical region guided by the textual prompt $\nu^{\Lambda}_{k}$ and the visual prompt $\nu^{I}_{k}$, where the image $I$ is inpainted by a spatial grid. Then, two validators subsequently ensure that the proposed region is both collision-free and semantically consistent with the instruction $\Lambda$. Note that the object identification module runs once beforehand to generate object masks used during the validation process.
    }
    \label{fig:method}
\end{figure*}

As VLMs demonstrate generalization across diverse tasks~\cite{brohan2023can,shah2023lm}, researchers have begun applying them to spatial grounding~\cite{yuan2025robopoint}. These models enable direct prediction of goal points grounded in visual scenes from natural language instructions. RoboPoint~\cite{yuan2025robopoint}, for example, fine-tunes VLMs on spatial phrases to predict 2D keypoints, translating relational commands into precise spatial points. Recent VLMs, such as Molmo~\cite{deitke2025molmo} and Gemini 2.5~\cite{comanici2025gemini}, further demonstrate zero-shot point prediction from language instructions. However, their point-based outputs remain coarse and lack fine-grained spatial precision. Our method enhances this limitation by introducing a coarse-to-fine inference framework that returns fine-grained regions for robotic task executions.

\noindent\textbf{Space representation}: Space representation plays a crucial role at the intersection of spatial grounding and robotic task execution. Early robotic studies model space using discrete grid or volumetric maps to capture the geometry of free or occupied regions~\cite{elfes2002using, thrun2003learning}. Researchers then introduce continuous representations, such as Euclidean or truncated signed distance fields to capture smoother geometric relationships~\cite{oleynikova2016signed,oleynikova2017voxblox}. Recent studies augment these field representations with object-level semantics through semantic mapping~\cite{grinvald2019volumetric,garg2020semantics}, although precise associations near region boundaries remain challenging. Alternatively, researchers employ superpixelization~\cite{achanta2010slic} to group homogeneous pixels into region-based representations that preserve sharp regional boundaries and coherent characteristics~\cite{zhu2023superpixel,yang2024rgb}. Building on this idea, our method introduces learning-based superpixelization that refines VLM-generated coarse predictions in accordance with the surrounding environmental context for accurate robotic perception and mapping.
\section{Methodology}
We introduce \method, a hierarchical space-grounding method that combines grid-guided VLM prompting with superpixel-based refinement.

\subsection{Overview}
Consider an input instruction $\Lambda$ and an input RGB image $I \in \mathbb{Z}^{3\times H \times W}$ containing $N$ objects, where $H$ and $W$ denote the image height and width, respectively. Our goal is to predict a space mask (i.e., segment) $\mathcal{M}_\Lambda \in \mathbb{Z}^{H \times W}$ that corresponds to $\Lambda$. As shown in Fig.~\ref{fig:method} (top), inference proceeds in two steps: 1) \textit{space reasoning}, and 2) \textit{space refinement}.
This modular design improves grounding accuracy and reliability while mitigating hallucinations in the VLM. 

In detail, the \textit{space-reasoning} step infers a canonical region $\tilde{\mathcal{M}}_\Lambda\in \mathbb{Z}^{H\times W}$ leveraging a grid-guided visual-text prompt, and iteratively refines it until the region satisfies both physical and semantic constraints with respect to $\mathcal{M}_{o_{1:N}}$ and $\Lambda$. Finally, the \textit{space-refinement} step locally adapts the canonical region to precisely fit the environment using superpixels. 

The grid guidance helps the VLM distinguish low or texture free space lacking distinctive features while maintaining semantic consistency with the instruction. Superpixel-based refinement reduces computational cost and enhances alignment with spatio-semantic pixel distribution compared to pixel-level refinement. We describe each component in detail below.

\subsection{Grid-guided \textit{Space Reasoning}}
This step guides the VLM to propose a canonical region $\tilde{\mathcal{M}}_\Lambda$ maximizing its reasoning capability. Fig.~\ref{fig:method} (bottom) illustrates the iterative reasoning and validation process. At each iteration, the prompt generator creates a visual prompt $\mathbf{\nu}^I$ for grid-based guidance and a text prompt $\mathbf{\nu}^T$ to interpret the instruction $\Lambda$. Feeding these concatenated prompts into the VLM yields $M$ ellipses $[\varepsilon_1, ..., \varepsilon_M]$, where each $\varepsilon_j$ represents a proposed region. Note that $M$ typically ranges from one to two depending on the VLM output. Then, we combine ellipses to form the predicted region $\tilde{\mathcal{M}}_\Lambda$. 

To validate $\tilde{\mathcal{M}}_\Lambda$, we introduce two validators: a physical validator to assess feasibility for object placement, and a semantic validator to ensure consistency with the instruction $\Lambda$. We detail each component below. 

\noindent\textbf{1) Object identification} 
Prior to grounding, we construct a joint object mask $\mathcal{M}_{o_{1:N}}=\mathcal{M}_{o_1}\cup ... \cup \mathcal{M}_{o_N} $, where each $\mathcal{M}_{o_i}\in\mathbb{Z}^{H \times W}$ denotes the binary mask of the $i$-th object. We use the constructed mask in the validation process of the \textit{space‑reasoning} step. As our focus is on space grounding without prior object knowledge, we employ Grounded-SAM~\cite{ren2024grounded}, an open-set object-mask identifier that first detects object bounding boxes using Grounding DINO~\cite{liu2023grounding}, and then extracts the corresponding masks using SAM~\cite{kirillov2023segment}, conditioned on the detected boxes.

\noindent\textbf{2) Prompt generator}: At each iteration $k$, our generator produces a novel grid-guided visual prompts $\mathbf{\nu}^I_k$ by overlaying a grid $I^{\text{grid}}\in\mathbb{Z}^{3\times H\times W}$ onto the input image $I$, providing explicit visual cues. We draw the grid $I^{\text{grid}}$ in black with a thickness of $1.4$ pixels at $100$ DPI and $100$-pixel intervals, regardless of the size of the image. In the grid, we also display axis tick values and labels (e.g., ``x axis'') to support reasoning about direction and distance. The grid remains on the top layer throughout all iterations. We define the initial prompt as $\mathcal{\nu}^{I}_0=I\oplus I^{\text{grid}}$, where $\oplus$ denotes the overlay operation. From the second iteration ($k>0$), we additionally overlay the latest predicted region $\tilde{\mathcal{M}}_\Lambda$ as red pixels: $ \mathcal{\nu}^{I}_{k>0}  =
  I \oplus I^\text{grid} \oplus \tilde{\mathcal{M}}_\Lambda$. 

Alongside, the generator produces a text prompt $\mathbf{\nu}^\Lambda_k$ using an LLM to guide the VLM in decomposing the grounding process while interpreting the visual prompt $\mathbf{\nu}^I_k$. The prompt includes (i) \textit{object guidance}\textemdash identifying instruction-relevant objects and their spatial extent based on $\Lambda$, (ii) \textit{region guidance}\textemdash prompting the VLM to output region coordinates, and (iii) \textit{placement feasibility guidance}\textemdash —ensuring that the predicted region is feasible for object placement as shown in Fig.~\ref{fig:prompt_1}.
From iteration $k>0$, the prompt also incorporates the feedback from the validators, enabling the VLM to refine proposals based on prior errors. Note that the VLM may return coordinates for multiple ellipses.

\vspace{1em}
\begin{promptblock}
Generate the center coordinates and semi-axis lengths of an elliptical region within an image, based on the provided overhead image and spatial instructions. The regions should be compact, accurate, and feasible to place.\\
...\\
\# Steps\\
1. **Extract Information**: Identify relevant objects and their extents from the image and spatial instructions.\\
2. **Generate Regions**: Determine suitable coordinates for elliptical regions following the given spatial instructions.\\
3. **Ensure Feasible Placement**: Verify that the generated coordinates are placed in valid, unobstructed locations according to the feasibility principles.\\

\# Output Format\\
The ``center\_coordinates" should be a list of center coordinates in the form of [[X1, Y1]].\\
The ``semi\_axes\_lengths" should be the lengths of semi-major and semi-minor axes in the form of [[a, b]], a $>=$ b.\\
The``angle" should be the tilted angle of the ellipse in degrees.\\
Spatial instruction: ...
\end{promptblock}
\captionof{figure}{A capture of the region proposal prompt for $k=0$}
\vspace{1em}
\label{fig:prompt_1}

\noindent\textbf{3) VLM-based region proposer}: Upon receiving the two prompts $\mathcal{\nu}^{\mathbf{I}}_k$ and $\mathcal{\nu}^\Lambda_k$, the VLM predicts a unified region $\tilde{\mathcal{M}}_\Lambda\in\mathbb{Z}^{H\times W}$ consisting of canonical region proposals, represented as ellipses. We parameterize each ellipse $\varepsilon_j$ using its center coordinates, semi-axis lengths, and rotation angle, extracted directly from the VLM's structured output via a text-to-ellipse conversion. Given these parameters, we generate individual elliptical masks and combine them to form a final region $\tilde{\mathcal{M}}_\Lambda$ through logical union.

\noindent\textbf{4) Physical \& semantic validators}: To ensure that the proposed region $\tilde{\mathcal{M}}_\Lambda$ satisfies both the physical and semantic requirements of the instruction $\Lambda$, we conduct a two-stage validation at each iteration. The first stage checks the physical validity of the proposed region mask. In the case where $\tilde{\mathcal{M}}_\Lambda$ intersects with the joint object mask $\mathcal{M}_{o_{1:N}}$, we further assess whether the intersection supports valid placements (e.g., ``on the dish'' or ``in the basket''). Otherwise, we regard $\tilde{\mathcal{M}}_\Lambda$ is suitable for placement actions.

For the further assessment, we issue another VLM query with a validation prompt consisting of a visual prompt $\mathcal{\nu}^{I, \text{phy}}_k=
I \oplus I^\text{grid} \oplus \tilde{\mathcal{M}}_\Lambda$ and a text prompt $\mathcal{\nu}^{\Lambda, \text{phy}}_k$ that asks whether placing an object at $\tilde{\mathcal{M}}_\Lambda$ is physically feasible, yielding a binary response as shown in Fig.~\ref{fig:prompt_3}. 

\vspace{1em}
\begin{promptblock}
You are a placement feasibility validator. You will be provided with a spatial instruction and an image containing a red elliptical region. Analyze the image and answer the following questions about the red elliptical region, providing detailed reasoning for each response, followed by a pass/fail conclusion for each question.\\
...\\
\# Judgment Guidelines\\
- Describe your reasoning for each question in detail, then make a judgment.\\
- Do not include any suggestions for improvement in your response.\\

\# Questions to be answered:
1. **Placement Feasibility Check**: Find out objects overlap with the predicted region. If overlap with the objects is permitted, then answer ``pass''; otherwise, answer ``fail''.\\

Answer each question by referencing the image and spatial instructions. After your reasoning for each question, include a ``pass'' or ``fail'' judgment for clear assessment.
\end{promptblock}
\captionof{figure}{A capture of the physical validator prompt}
\vspace{1em}
\label{fig:prompt_3}

For semantic validation, we reuse the visual prompt $\mathcal{\nu}^{I, \text{sem}}_k$ ($=\mathcal{\nu}^{I, \text{phy}}_k$) and provide a semantic text prompt $\mathcal{\nu}^{\Lambda, \text{sem}}_k$, asking whether $\tilde{\mathcal{M}}_\Lambda$ satisfies the spatial semantics of $\Lambda$ as in Fig.~\ref{fig:prompt_4}. To improve accuracy, we instruct the VLM to decompose compositional instructions and validate each sub-component within $\mathcal{\nu}^{\Lambda, \text{sem}}_k$. If $\tilde{\mathcal{M}}_\Lambda$ passes both validations or if the process reaches the maximum number of iterations, we return it; otherwise, we return to the prompt generation step.

\vspace{1em}
\begin{promptblock}
You are a spatial feasibility validator. You will be provided with a spatial instruction and an image containing a red elliptical region. Analyze the image and answer the following questions about the region, providing detailed reasoning for each response, followed by a pass/fail conclusion for each question.\\
The region has already gone through a placement feasibility test and it has been verified that it is feasible to place and does not overlap with any object it should avoid.\\
...\\
\# Judgment Guidelines\\
- Describe your reasoning for each question in detail, then make a judgment.\\
- Do not include any suggestions for improvement in your response.\\

\# How to check if the red region is satisfying the instruction:\\
1) **Break the instruction into multiple segments.**\\
  - Example 1:\\
    - Instruction: ``To the right of the tray with edible items and below the spectacles.''\\
    - Segments: [right of the tray with edible items, below the spectacles]\\

2) **Spatial validation of the red elliptical region for each segment:** For each segment, check whether the position of the red elliptical region satisfies it. Use the spatial guidelines to validate the position of the region for each segment. ...\\

3) If the object mentioned in the instruction has multiple instances in the image, carefully verify that the red elliptical region follows the spatial instruction relative to the correct instance of that object.\\

\# Questions to be answered:\\
1. **Spatial Compliance**: Determine whether the region visually satisfies the spatial instruction provided. Describe your reasoning. Use the ``spatial guidelines'' and ``how to check if red region is satisfying the instruction'' to reach a conclusion.\\

Answer each question by referencing the image and spatial instruction. After your reasoning for each question, include a ``pass'' or ``fail'' judgment for clear assessment.
\end{promptblock}
\captionof{figure}{A capture of the semantic validator prompt}
\vspace{1em}
\label{fig:prompt_4}

\begin{figure*}[t]
    \centering
    \includegraphics[width=\linewidth]{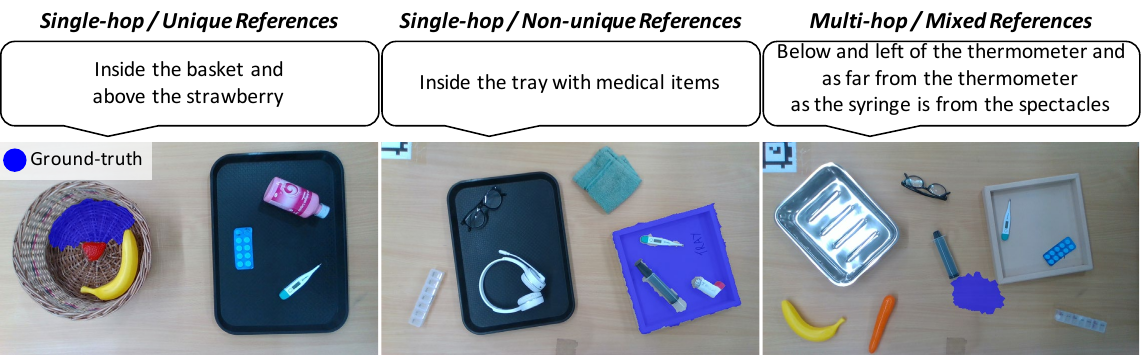}
    \caption{
    Examples in the proposed space-grounding benchmark. Each example includes an instruction-image pair with the corresponding ground-truth space label (blue). For labeling consistency and efficiency, human annotators select superpixel-based regions directly on the image.  }
    \label{fig:setup}
\end{figure*}

\subsection{Superpixel-based \textit{space refinement}}
We locally adapt the coarse canonical region $\tilde{\mathcal{M}}_\Lambda$ to the fine-grained structure of the surrounding environment as well as free space, we predict the final manipulation region $\mathcal{M}_\Lambda$ by modeling its residual. We particularly introduce a residual learning module by decomposing the instructed region as $\mathcal{M}_\Lambda = \tilde{\mathcal{M}}_\Lambda \oplus \mathcal{M}_\Lambda^{\text{residual}}$ where $\mathcal{M}_\Lambda^{\text{residual}}$ captures local refinements over the superpixel space. To simplify learning, we approximate this decomposition in the logit space as
\begin{align}
l_{\Lambda}=\alpha \tilde{l}_{\Lambda}+(1-\alpha)l_{\Lambda}^{\text{residual}},
\label{eq_logit}
\end{align}
where $l_\Lambda$, $\tilde{l}_\Lambda$, and $l_\Lambda^{\text{residual}}$ denote the superpixel-wise logits of $\mathcal{M}_\Lambda$, $\tilde{\mathcal{M}}_\Lambda$, and $\mathcal{M}_\Lambda^{\text{residual}}$, respectively; $\alpha\in[0,1]$ is a scaling factor and $|l_\Lambda|=L$ with $L$ denoting the number of superpixels. We describe superpixel generation and logit estimation below.

To compute $\tilde{l}_\Lambda$ for the predicted region $\tilde{\mathcal{M}}_\Lambda$, we generate superpixels from the grayscale image $I^{\text{gray}}\in\mathbb{R}^{H\times W}$ using SLIC~\cite{achanta2010slic}. We then assign a pseudo logit to each superpixel in two steps:
\begin{align}
\tilde{\mathcal{M}}_\Lambda \xrightarrow{smoothing}\tilde{\mathcal{M}}'_\Lambda \xrightarrow{aggregation} \tilde{l}_\Lambda,
\end{align}
where $\tilde{\mathcal{M}}'_\Lambda$ represents a center-distance weighted value for each ellipse; pixels farther from the center of each ellipse have smaller values. We then compute the pseudo-logit value $\tilde{l}_\Lambda$ by averaging the pixel values within each superpixel.

To model the residual $l_{\Lambda}^{\text{residual}}$, we construct a superpixel graph where each node represents a superpixel and each edge connects adjacent superpixels. Node features consist of the mean, minimum, and maximum values within each superpixel in $I^{\text{gray}}$ and $\tilde{\mathcal{M}}_{\Lambda}$, along with a binary indicator specifying whether the instruction $\Lambda$ requires distance reasoning, identified by the LLM. 

We use a graph neural network, GPS~\cite{rampavsek2022recipe}, to predict the superpixel-wise residual logit $l_{\Lambda}^{\text{residual}}$. To supervise it, we compute a focal loss of the predicted probabilities $\mathbf{p}=\sigma(l_\Lambda)\in\mathbb{R}^L$ given the ground-truth space labels, where $\sigma$ is the sigmoid function. Finally, we project the superpixel-wise probabilities $\mathbf{p}$ back to the pixel space and binarize it, resulting in the final refined region $\mathcal{M}_\Lambda$.
\section{Experimental Setup}
Our experiments aim to measure performance improvements in space grounding tasks that require reasoning.

\subsection{Benchmark description}
We introduce a superpixel-level space grounding benchmark consisting of real-world scene images, natural language instructions in English, and human-annotated ground-truth labels.
We capture tabletop scenes that may include containers holding other objects, as well as multiple identical items.
The instructions cover nine types of spatial relations: `left,' `right,' `above,' `below,' `near,' `far,' `inside,' `outside,' and `along the direction of.' We annotate the ground-truth labels at the superpixel level rather than the pixel level. We collected the data with approval from the institutional review board (IRB).

We categorize the instructions into three types based on the number of reasoning hops and the uniqueness of reference objects, as illustrated in Fig.~\ref{fig:setup}:
(i) \textit{single-hop space reasoning with unique references},
(ii) \textit{single-hop space reasoning with non-unique references}, and 
(iii) \textit{multi-hop space reasoning with unique/non-unique references}

The first category, \textit{single-hop space reasoning with unique references}, includes instructions containing one to four spatial expressions with clearly identifiable reference objects. These cases require only direct interpretation of spatial terms and localization of the reference objects, as illustrated by the example ``inside the basket and above the strawberry'' in the first column of Fig.~\ref{fig:setup}.

The second category, \textit{single-hop space reasoning with non-unique references}, includes instructions that require resolving ambiguous references. These ambiguities often arise in scenes with multiple visually similar objects, as shown in the middle column of Fig.~\ref{fig:setup}.

The third category, \textit{multi-hop space reasoning with unique/non-unique references}, involves multi-hop inference. The model needs to understand the distance between two objects and apply that distance relative to another reference object, often requiring multiplicative reasoning. For example, the instruction ``Above the inhaler by twice the distance between the inhaler and the strawberry,''
belongs to the \textit{multi-hop spatial reasoning} category.

The benchmark contains a total of $350$ samples, which are split into $200$ for training, $50$ for validation, and $100$ for testing. The validation and test sets maintain a balanced distribution across the three instruction categories by design.

\begin{figure*}[t]
    \centering
    \includegraphics[width=\linewidth]{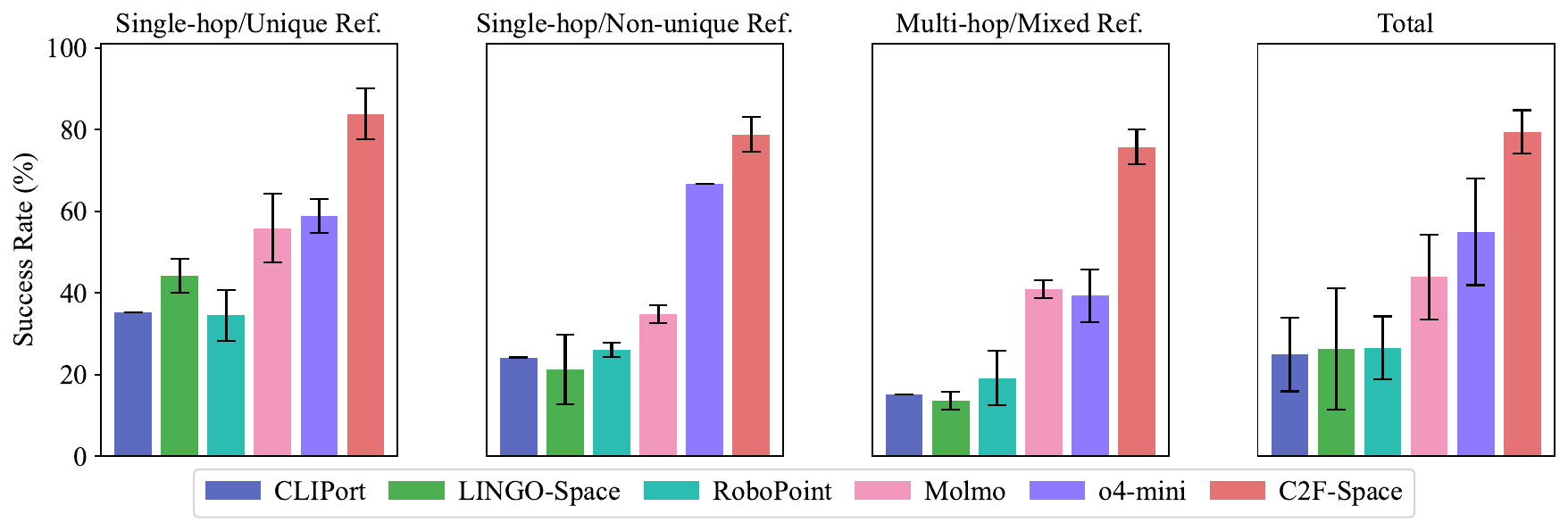}
    \caption{Performance comparison in terms of success rate [$\%$]. The `Total' denotes the average across three instruction categories. The error bar denotes standard deviation.
    }
    \label{fig:result}
\end{figure*}

We evaluate the performance of our method and baselines using two metrics. The first is the \textit{success rate}, which considers a prediction successful if either the maximum-probability point or the centroid of the predicted mask lies within the ground-truth region.
However, the success rate is a binary indicator and therefore cannot capture the quality of the predicted mask in finer detail. For example, even if the predicted mask covers only a small portion of the ground-truth area, the success rate remains the same as long as the maximum-probability point falls inside the ground truth.
To address this limitation, we additionally measure the \textit{intersection over union (IoU)} between the predicted and ground-truth masks in pixel space, which reflects how well the predicted region overlaps with the true target area.

\subsection{Baseline Methods}
We evaluate five space-grounding baselines:
\begin{itemize}[leftmargin=*, noitemsep,topsep=0pt]
    \item \textbf{\textsc{CLIPort}}~\cite{shridhar2022cliport}: A language-conditioned imitation learning approach that predicts a pixel-level target location for pick-and-place manipulation using CLIP-based~\cite{radford2021learning} image-language encoding. We disable rotational augmentation and modify the loss function to incorporate a mask as the ground-truth target. 
    \item \textbf{LINGO-Space}~\cite{kim2024lingo}: A probabilistic space-grounding method that incrementally estimates the spatial distribution of target regions from composite referring expressions using configurable polar distributions. We modify the loss function to incorporate a mask as the ground-truth target instead of a single point.
    \item \textbf{\textsc{RoboPoint}}~\cite{yuan2025robopoint}: A VLM fine-tuned with synthetic instructions to predict keypoint affordances from language, including spatial descriptions. \textsc{RoboPoint} predicts multiple keypoints for space grounding. We use its pre-trained model and evaluate performance by computing the success rate, defined as the ratio of predicted points within the ground-truth mask. Note that we do not measure IoU for \textsc{RoboPoint}.
    \item Molmo 72B~\cite{deitke2025molmo}: An open-source VLM designed for multimodal reasoning, capable of 2D pointing to localize and reference specific regions within images. We adopt the point-based prompting strategy from \textsc{RoboPoint} and do not measure IoU as \textsc{RoboPoint}. 
    \item \textbf{GPT-o4} mini~\cite{openai2025o4}: A GPT-o4 mini with a point-based prediction prompt as \textsc{RoboPoint} and \textsc{Molmo}.
\end{itemize}

We apply Otsu's thresholding to convert a predicted probability into a binary mask for both our method and baselines, except RoboPoint, since spatial regions tend to exhibit low confidence, making it difficult to select a universal threshold across different cases. We repeat the experiment twice for each method.

\subsection{Implementation details}
For object identification, we use Grounding DINO-B~\cite{liu2023grounding}, which combines a SWIN-B~\cite{liu2021swin} visual backbone with a BERT-base-uncased~\cite{devlin2019bert} text backbone, together with SAM employing a ViT-B~\cite{vit} backbone. We prompt Grounding DINO with the query `object, objects' to extract bounding boxes and subsequently use these boxes to prompt SAM. For grid-guided space reasoning, we employ the `o4-mini-2025-04-16' model across all tasks. We limit the maximum number of iterations to two, as additional iterations do not yield noticeable performance improvements in our experiments. During evaluation, we enable automatic retries when API-related errors occur, such as missing parsing results.

\begin{figure*}[t]
    \centering
    \includegraphics[width=\linewidth]{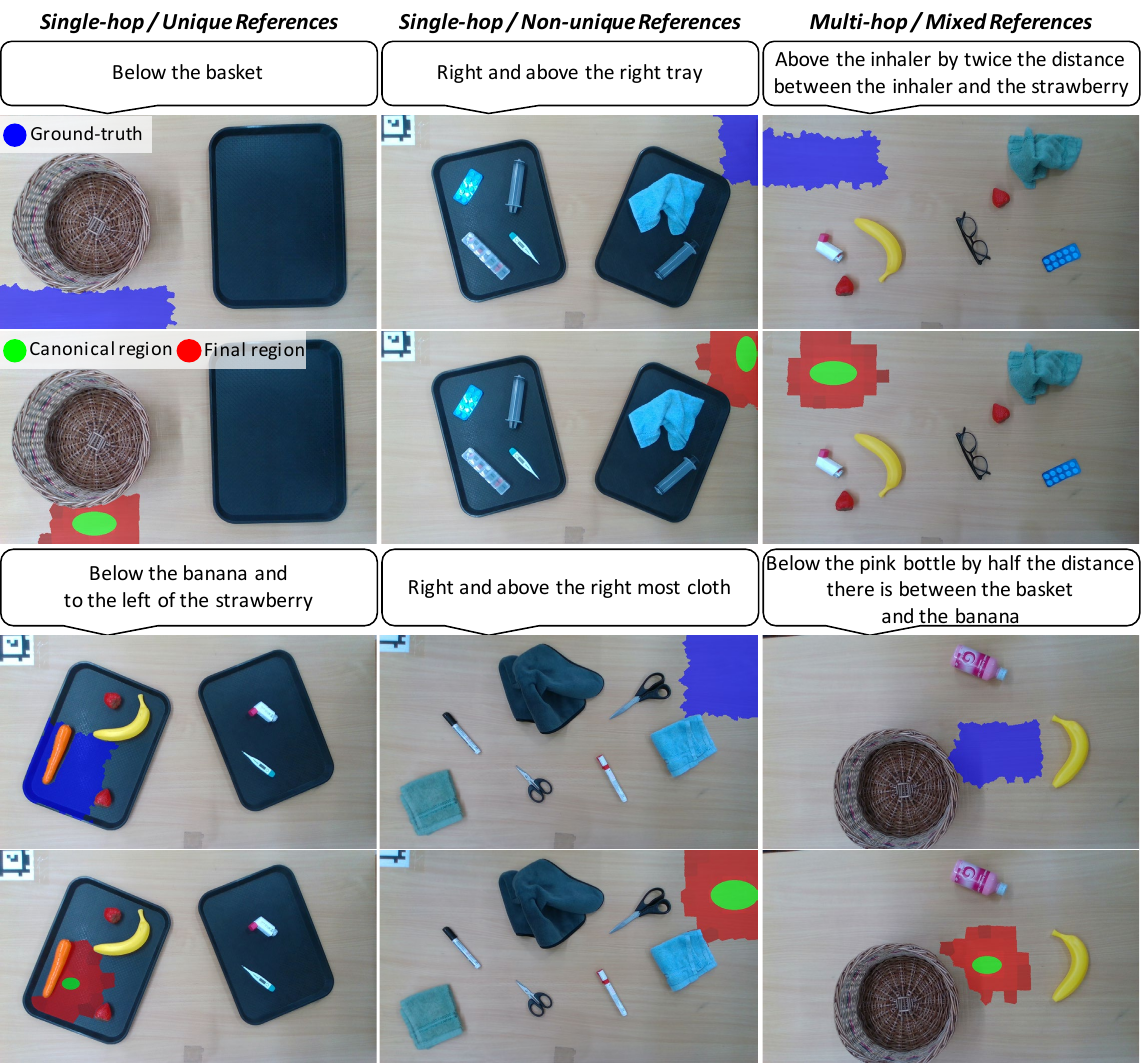}
    \caption{
    Demonstrations of \method\ predictions across six space-grounding scenarios. For each linguistic instruction, the region proposer generates a canonical region (green), while the refinement module adapts it to the surrounding environment without over-relaxation using superpixelization (red). The blue regions indicate ground-truth regions. Note that we measure the distance between the centers of two referenced objects. }
    \label{fig:res}
\end{figure*}
\section{Evaluation}
\begin{figure}
    \centering
    \includegraphics[width=\linewidth]{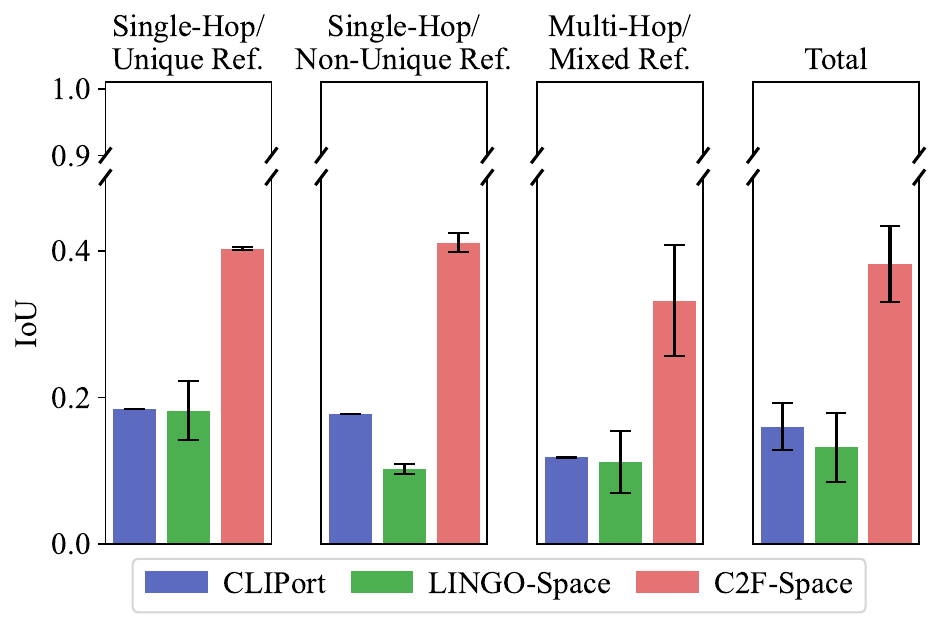}
    \caption{Performance comparison in terms Intersection over Union (IoU). The `Total' denotes the average across three instruction categories. The error bar denotes standard deviation. }
    \label{fig:iou}
\end{figure}

\textbf{Benchmark evaluation.}
We evaluate our method and baselines on the proposed space-grounding benchmark. As shown in Fig.~\ref{fig:result}-~\ref{fig:iou}, our approach consistently outperforms all baselines in both success rate and IoU across all instruction categories. Although built on o4-mini, the gains over the o4-mini backbone indicate that grid-based visual prompting and superpixel-based refinement substantially strengthen grounding performance. While large-scale VLMs (i.e., Molmo and o4-mini) achieve similarly high success rates exceeding $55\%$ on \textit{single-hop space reasoning with unique references} task, Molmo's performance drops markedly with non-unique references. This indicates o4-mini maintains stable performance by leveraging stronger reasoning to infer implicit contextual cues that resolving referential ambiguity requires robust reasoning capabilities. These findings justify our choice of o4-mini as the foundation of the \textit{spatial reasoning} module. 

In contrast, small models exhibit poor performance due to limited reasoning capacity. The small-scale VLM, \textsc{RoboPoint}, achieves significantly lower success rates, not only due to its weak reasoning ability but also since it learns from a predefined set of spatial relations, failing to generalize to unseen concepts such as `far.' Similarly, other learning-based models, \textsc{CLIPort} and LINGO-Space, also show lower success rates due to limited generalization to novel relations and objects, as they rely on \textsc{CLIP}'s limited visual–textual representations.

The results for \textit{multi-hop space reasoning} particularly demonstrate the effectiveness of the grid-inpainting reasoning and propose-validation loop in \method. While \method achieves approximately $76\%$ success, all baselines fall below $50\%$, as this task require understanding inter-object relationships and contextual dependencies. The grid overlaid on the image provides an explicit spatial structure, enabling o4-mini to estimate the relative distances between objects and free spaces. Moreover, the validation loop in \method iteratively refines prior predictions using visual feedback. Together, these modules establish a robust multi-hop spatial reasoning mechanism that enables our method to succeed in tasks where other baselines fail.

Fig.~\ref{fig:iou} shows \method's superior space generation performance, achieving an average IoU of $0.382$ across categories—more than twice that of the second-best baseline, \textsc{CLIPort} ($0.160$). This high IoU results not only from semantically accurate region predictions but also from fine-grained superpixelization. As shown in the first and second columns of Fig.~\ref{fig:res}, the refinement module effectively captures object boundaries, enhancing spatial precision. Moreover, as space grounding inherently lacks well-defined object boundaries, the IoU may remain relatively low (e.g., $0.493$), even when the grounded region is qualitatively reasonable, as demonstrated in the first column of Fig.~\ref{fig:res}.

\begin{figure}[t]
    \centering
    \includegraphics[width=\linewidth]{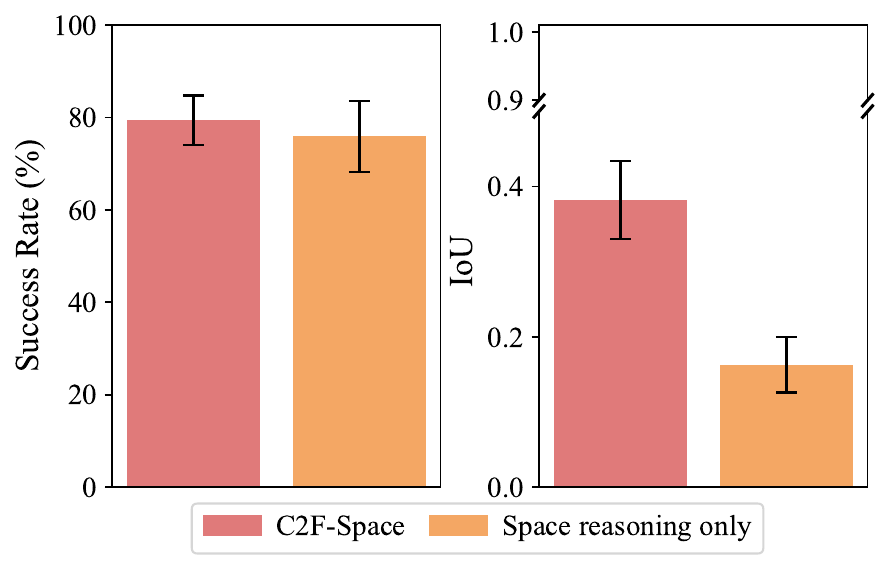}
    \caption{
    Ablation study evaluating the effectiveness of the superpixel-based space refinement module in \method. We compare \method\ with `space reasoning only' variant (i.e., without the refinement module) in terms of success rate [$\%$] and IoU.
    }
    \label{fig:ab0}
\end{figure}
\begin{figure}[t]
    \centering
    \includegraphics[width=\linewidth]{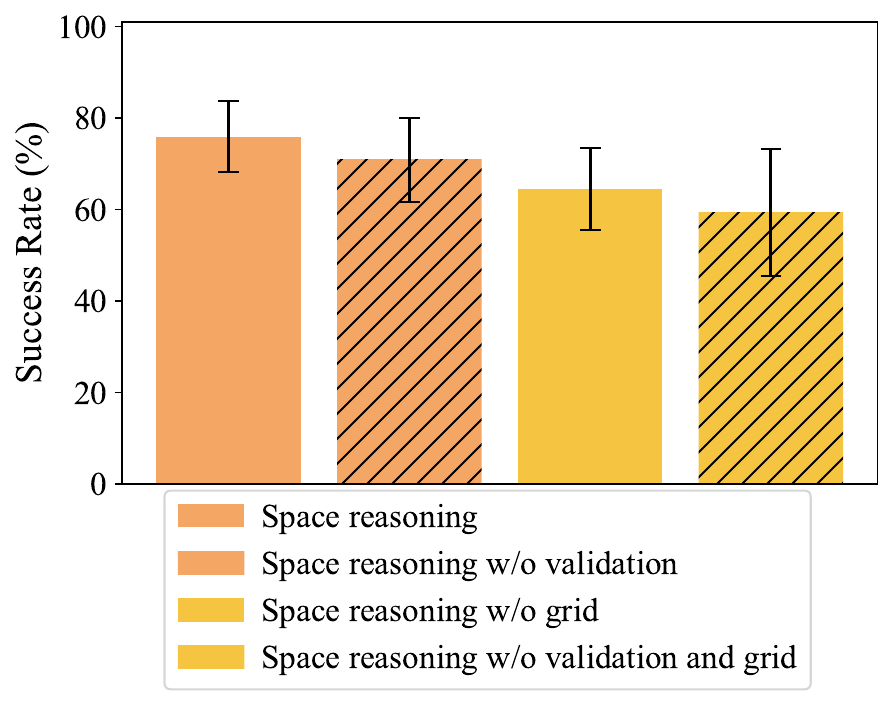}
    \caption{
    Ablation study evaluating the effectiveness of the grid-guided \textit{space reasoning} module. We assess performance degradation when removing either the validation component, the grid-based visual prompts, or both.}
    \label{fig:ab1}
\end{figure}

\begin{figure*}[t]
    \centering
    \includegraphics[width=\textwidth]{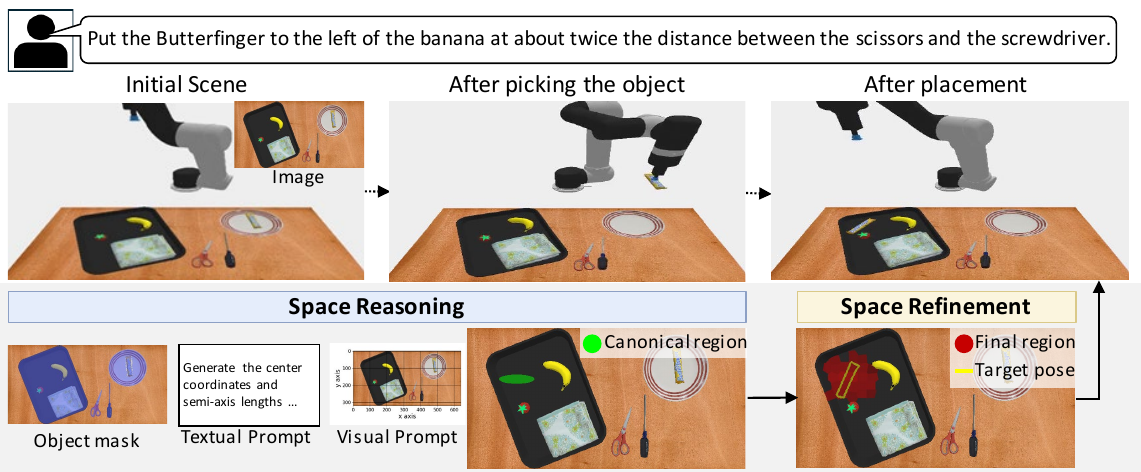}
    \caption{A pick and place task demonstration in PyBullet simulator. Given the \textit{multi-hop} instruction and the top-down view scene image, \method\ indentifies space occupied by objects, reasons space using VLMs, and then refines the generated region. Note that we measure the distance between the centers of the two referenced cloths.}
    \label{fig:demo}
\end{figure*}

\textbf{Ablation studies.}
We conduct ablation studies to evaluate the impact of superpixel-level refinement. The success rate remains comparable with or without this refinement, as shown in Fig.~\ref{fig:ab0}, indicating that reasoning is primarily based on the VLM module. Including the superpixel-level refinement improves IoU by more than twice, resulting in the highest IoU among all baselines. As shown in the bottom rows of Fig.~\ref{fig:res}, the refinement module adjusts the prediction without overlapping objects.

We further analyze the contribution of each component in the grid-guided space reasoning module, focusing on the grid-based visual prompt and the validation step.
As shown in Fig.~\ref{fig:ab1}, removing the validation step and its corresponding iteration results in a slight performance drop of approximately $5\%$ in success rate. In contrast, removing the grid-based visual prompt causes a substantial decrease of $11.5\%$, and eliminating both components leads to an additional $5.0\%$ decline. These findings underscore the crucial role of grid-guided visual prompting in enabling effective space grounding, as it provides essential spatial structure for the VLM's reasoning process.

\textbf{Simulation demonstrations.}
We finally demonstrate the applicability of \method\ to robotic manipulation tasks using the PyBullet-based simulator with UR5 manipulator from \textsc{CLIPort} \cite{shridhar2022cliport}. Fig.~\ref{fig:demo} presents a demonstration sequence along with the intermediate outputs of each module, given a human instruction and a top-down image. 

First, the \textit{space-reasoning} module predicts a canonical region (green) by reasoning through o4-mini. The subsequent \textit{space-refinement} module then produces fine-grained superpixels (red) with associated prediction scores above a certain threshold. We use the predicted superpixels as a spatial mask to sample the optimal placement position of the target object, maximizing the intersection-weighted score sum within the mask. 

After identifying the target object and placement position, the simulated robot grasps the object specified in the instruction and places it at the predicted location using a predefined action primitive. As shown in Fig.~\ref{fig:demo}, the robot successfully positions the target object in the instructed spatial region.
\section{Conclusion}
We proposed \method, a two-stage space-grounding framework that combines a VLM with a superpixel-level refinement module to identify regions that are physically feasible and semantically consistent with the input instruction. Our method first uses the VLM to globally predict a coarse region based on the novel \textit{grid-inpainting} prompting technique. Then, it locally refines this coarse prediction into a precise region using superpixel-level decisions. This design allows for accurate localization and robust generalization across complex spatial expressions. Experimental results on our new benchmark show that superpixel-based refinement significantly improves IoU without reducing the success rate. In addition, the \textit{grid-inpainting} prompt with the propose-validate strategy plays a key role in guiding the prediction process. Overall, \method\ outperforms existing baselines in grounding accuracy, demonstrating its effectiveness in resolving spatial references within complex scenes and instructions. We finally demonstrate the practical applicability of \method\ in robotic manipulation tasks through simulation, where the robot successfully places objects according to spatial instructions.
\bmhead{Acknowledgements}
This work was supported by the Institute of Information \& communications Technology Planning \& Evaluation (IITP) grant funded by the Korea Technology (MSIT) (No. RS-2022-II220311 and RS-2024-00336738), by the National Research Council of Science \& Technology (NST) grant by the MSIT (No. GTL25041-000),
Artificial intelligence industrial convergence cluster development project funded by the MSIT \& Gwangju Metropolitan City. 
We thank Mr. Mohd. Nadir, Mr. Aman Tambi, Mr. Sandeep Zachariah and Mr. Ribhu Vaypei for contributing towards the Franka Emika Robot test bed setup, data collection/processing and system development/CAD modeling support during the early phase of the project.  
Rohan Paul acknowledges research grant support from the ICMR-National Center for Assistive Health Technology (NCAHT), Govt. of India as project: RP04267 under School of IT (SIT), IIT Delhi, Hardware support from the Equipment Matching Grant from Industrial Research \& Development (IRD) Unit at IIT Delhi and administrative support from Mr. Shailendra Negi and Ms. Sunita Negi.  

\bibliography{sn-bibliography}

@string{icra = {Proceedings of the IEEE International Conference on Robotics and Automation (ICRA)}}

@string{corl = {Proceedings of the Conference on Robot Learning (CoRL)}}

@string{iros = {Proceedings of the IEEE/RSJ International Conference on Intelligent Robots and Systems (IROS)}}

@string{icml="Proceedings of the International Conference on Machine Learning (ICML)"}

@string{rss="Proceedings of Robotics: Science and Systems (RSS)"}

@string{aaai="Proceedings of the National Conference on Artificial Intelligence (AAAI)"}

@string{neurips="Conference on Neural Information Processing Systems (NeurIPS)"}

@string{iclr = {Proceedings of the International Conference on Learning Representation (ICLR)}}

@string{iccv="Proceedings of the International Conference on Computer Vision (ICCV)"}

@string{cvpr="Proceedings of the IEEE/CVF conference on computer vision and pattern recognition"}

@string{ral = "IEEE Robotics and Automation Letters (RA-L)"}

@string{eccv="Proceedings of the European Conference on Computer Vision (ECCV)"}

@string{tpami="IEEE transactions on pattern analysis and machine intelligence(TPAMI)"}

@string{emnlp="Proceedings of the Conference on Empirical Methods in Natural Language Processing (EMNLP)"}

@string{naacl="Proceedings of the Conference of the North American Chapter of the Association for Computational Linguistics (NAACL)"}

@inproceedings{deitke2025molmo,
  title={Molmo and pixmo: Open weights and open data for state-of-the-art vision-language models},
  author={Deitke, Matt and Clark, Christopher and Lee, Sangho and Tripathi, Rohun and Yang, Yue and Park, Jae Sung and Salehi, Mohammadreza and Muennighoff, Niklas and Lo, Kyle and Soldaini, Luca and others},
  booktitle=cvpr,
  pages={91--104},
  year={2025}
}

@inproceedings{
vit,
title={An Image is Worth 16x16 Words: Transformers for Image Recognition at Scale},
author={Alexey Dosovitskiy and Lucas Beyer and Alexander Kolesnikov and Dirk Weissenborn and Xiaohua Zhai and Thomas Unterthiner and Mostafa Dehghani and Matthias Minderer and Georg Heigold and Sylvain Gelly and Jakob Uszkoreit and Neil Houlsby},
booktitle=iclr,
year={2021},
}

@inproceedings{devlin2019bert,
  title={Bert: Pre-training of deep bidirectional transformers for language understanding},
  author={Devlin, Jacob and Chang, Ming-Wei and Lee, Kenton and Toutanova, Kristina},
  booktitle=naacl,
  pages={4171--4186},
  year={2019}
}

@inproceedings{liu2021swin,
  title={Swin transformer: Hierarchical vision transformer using shifted windows},
  author={Liu, Ze and Lin, Yutong and Cao, Yue and Hu, Han and Wei, Yixuan and Zhang, Zheng and Lin, Stephen and Guo, Baining},
  booktitle=iccv,
  pages={10012--10022},
  year={2021}
}

@article{ren2024grounded,
  title={Grounded SAM: Assembling open-world models for diverse visual tasks},
  author={Ren, Tianhe and Liu, Shilong and Zeng, Ailing and Lin, Jing and Li, Kunchang and Cao, He and Chen, Jiayu and Huang, Xinyu and Chen, Yukang and Yan, Feng and others},
  journal={arXiv preprint arXiv:2401.14159},
  year={2024}
}

@inproceedings{radford2021learning,
  title={Learning transferable visual models from natural language supervision},
  author={Radford, Alec and Kim, Jong Wook and Hallacy, Chris and Ramesh, Aditya and Goh, Gabriel and Agarwal, Sandhini and Sastry, Girish and Askell, Amanda and Mishkin, Pamela and Clark, Jack and others},
  booktitle=icml,
  pages={8748--8763},
  year={2021},
  organization={PMLR}
}

@inproceedings{kim2024lingo,
  title={LINGO-Space: Language-conditioned incremental grounding for space},
  author={Kim, Dohyun and Oh, Nayoung and Hwang, Deokmin and Park, Daehyung},
  booktitle=aaai,
  volume={38},
  number={9},
  pages={10314--10322},
  year={2024}
}

@inproceedings{jain2023ground,
  title={Ground then Navigate: Language-guided Navigation in Dynamic Scenes},
  author={Jain, Kanishk and Chhangani, Varun and Tiwari, Amogh and Krishna, K Madhava and Gandhi, Vineet},
  booktitle=icra,
  pages={4113--4120},
  year={2023},
  organization={IEEE}
}

@article{comanici2025gemini,
  title={Gemini 2.5: Pushing the frontier with advanced reasoning, multimodality, long context, and next generation agentic capabilities},
  author={Comanici, Gheorghe and Bieber, Eric and Schaekermann, Mike and Pasupat, Ice and Sachdeva, Noveen and Dhillon, Inderjit and Blistein, Marcel and Ram, Ori and Zhang, Dan and Rosen, Evan and others},
  journal={arXiv preprint arXiv:2507.06261},
  year={2025}
}

@inproceedings{chen2022pali,
  title={Pali: A jointly-scaled multilingual language-image model},
  author={Chen, Xi and Wang, Xiao and Changpinyo, Soravit and Padlewski, Piotr and Salz, Daniel and Goodman, Sebastian and Grycner, Adam and Mustafa, Basil and Beyer, Lucas and others},
  booktitle=iclr,
  year={2023}
}

@inproceedings{li2023blip,
  title={Blip-2: Bootstrapping language-image pre-training with frozen image encoders and large language models},
  author={Li, Junnan and Li, Dongxu and Savarese, Silvio and Hoi, Steven},
  booktitle=icml,
  pages={19730--19742},
  year={2023},
  organization={PMLR}
}

@article{cheng2025pointarena,
  title={PointArena: Probing Multimodal Grounding Through Language-Guided Pointing},
  author={Cheng, Long and Duan, Jiafei and Wang, Yi Ru and Fang, Haoquan and Li, Boyang and Huang, Yushan and Wang, Elvis and Eftekhar, Ainaz and Lee, Jason and Yuan, Wentao and others},
  journal={arXiv preprint arXiv:2505.09990},
  year={2025}
}

@inproceedings{shridhar2022cliport,
  title={Cliport: What and where pathways for robotic manipulation},
  author={Shridhar, Mohit and Manuelli, Lucas and Fox, Dieter},
  booktitle=corl,
  pages={894--906},
  year={2022},
  organization={PMLR}
}

@inproceedings{zhao2023differentiable,
  title={Differentiable Parsing and Visual Grounding of Natural Language Instructions for Object Placement},
  author={Zhao, Zirui and Lee, Wee Sun and Hsu, David},
  booktitle=icra,
  pages={11546--11553},
  year={2023},
}

@inproceedings{gkanatsios2023energybased,
    title={Energy-based Models are Zero-Shot Planners for Compositional Scene Rearrangement},
    author={Gkanatsios, Nikolaos and Jain, Ayush and Xian, Zhou and Zhang, Yunchu and Atkeson, Christopher and Fragkiadaki, Katerina},
    booktitle=rss,
    year={2023}
}

@inproceedings{shah2023lm,
  title={Lm-nav: Robotic navigation with large pre-trained models of language, vision, and action},
  author={Shah, Dhruv and Osi{\'n}ski, B{\l}a{\.z}ej and Levine, Sergey and others},
  booktitle=corl,
  pages={492--504},
  year={2023},
  organization={PMLR}
}

@inproceedings{song2025robospatial,
  title={Robospatial: Teaching spatial understanding to 2d and 3d vision-language models for robotics},
  author={Song, Chan Hee and Blukis, Valts and Tremblay, Jonathan and Tyree, Stephen and Su, Yu and Birchfield, Stan},
  booktitle=cvpr,
  pages={15768--15780},
  year={2025}
}

@misc{openai2025o4,
    title={OpenAI o4‑mini (July 15 version)},
    author={OpenAI},
    year={2025},
    howpublished={\url{https://platform.openai.com/docs/models/o4-mini}},
}

@inproceedings{shtedritski2023does,
  title={What does clip know about a red circle? visual prompt engineering for vlms},
  author={Shtedritski, Aleksandar and Rupprecht, Christian and Vedaldi, Andrea},
  booktitle=iccv,
  pages={11987--11997},
  year={2023}
}

@inproceedings{cai2024vip,
  title={Vip-llava: Making large multimodal models understand arbitrary visual prompts},
  author={Cai, Mu and Liu, Haotian and Mustikovela, Siva Karthik and Meyer, Gregory P and Chai, Yuning and Park, Dennis and Lee, Yong Jae},
  booktitle=cvpr,
  pages={12914--12923},
  year={2024}
}

@article{zender2008conceptual,
  title={Conceptual spatial representations for indoor mobile robots},
  author={Zender, Hendrik and Mozos, O Mart{\'\i}nez and Jensfelt, Patric and Kruijff, G-JM and Burgard, Wolfram},
  journal={Robotics and Autonomous Systems},
  volume={56},
  number={6},
  pages={493--502},
  year={2008},
  publisher={Elsevier}
}

@article{skubic2004spatial,
  title={Spatial language for human-robot dialogs},
  author={Skubic, Marjorie and Perzanowski, Dennis and Blisard, Samuel and Schultz, Alan and Adams, William and Bugajska, Magda and Brock, Derek},
  journal={IEEE Transactions on Systems, Man, and Cybernetics, Part C (Applications and Reviews)},
  volume={34},
  number={2},
  pages={154--167},
  year={2004},
  publisher={IEEE}
}

@article{kulkarni2013babytalk,
  title={Babytalk: Understanding and generating simple image descriptions},
  author={Kulkarni, Girish and Premraj, Visruth and Ordonez, Vicente and Dhar, Sagnik and Li, Siming and Choi, Yejin and Berg, Alexander C and Berg, Tamara L},
  journal=tpami,
  volume={35},
  number={12},
  pages={2891--2903},
  year={2013},
  publisher={IEEE}
}

@inproceedings{lan2012image,
  title={Image retrieval with structured object queries using latent ranking svm},
  author={Lan, Tian and Yang, Weilong and Wang, Yang and Mori, Greg},
  booktitle=eccv,
  pages={129--142},
  year={2012},
  organization={Springer}
}

@inproceedings{lu2016visual,
  title={Visual relationship detection with language priors},
  author={Lu, Cewu and Krishna, Ranjay and Bernstein, Michael and Fei-Fei, Li},
  booktitle={European conference on computer vision},
  pages={852--869},
  year={2016},
  organization={Springer}
}

@inproceedings{stopp1994utilizing,
  title={Utilizing spatial relations for natural language access to an autonomous mobile robot},
  author={Stopp, Eva and Gapp, Klaus-Peter and Herzog, Gerd and Laengle, Thomas and Lueth, Tim C},
  booktitle={Proceedings of the German Annual Conference on Artificial Intelligence},
  pages={39--50},
  year={1994},
  organization={Springer}
}

@article{lu2019vilbert,
  title={Vilbert: Pretraining task-agnostic visiolinguistic representations for vision-and-language tasks},
  author={Lu, Jiasen and Batra, Dhruv and Parikh, Devi and Lee, Stefan},
  journal=neurips,
  volume={32},
  year={2019}
}

@inproceedings{tan2019lxmert,
  title={LXMERT: Learning Cross-Modality Encoder Representations from Transformers},
  author={Tan, Hao and Bansal, Mohit},
  booktitle=emnlp,
  pages={5100--5111},
  year={2019}
}

@inproceedings{tan2014grounding,
  title={Grounding spatial relations in natural language by fuzzy representation for human-robot interaction},
  author={Tan, Jiacheng and Ju, Zhaojie and Liu, Honghai},
  booktitle={IEEE International Conference on Fuzzy Systems (FUZZ-IEEE)},
  pages={1743--1750},
  year={2014}
}

@incollection{bloch2003representation,
  title={On the representation of fuzzy spatial relations in robot maps},
  author={Bloch, Isabelle and Saffiotti, Alessandro},
  booktitle={Intelligent systems for information processing},
  pages={47--57},
  year={2003},
  publisher={Elsevier}
}

@inproceedings{venkatesh2021spatial,
  title={Spatial reasoning from natural language instructions for robot manipulation},
  author={Venkatesh, Sagar Gubbi and Biswas, Anirban and Upadrashta, Raviteja and Srinivasan, Vikram and Talukdar, Partha and Amrutur, Bharadwaj},
  booktitle=icra,
  pages={11196--11202},
  year={2021}
}

@inproceedings{mees2020learning,
  title={Learning object placements for relational instructions by hallucinating scene representations},
  author={Mees, Oier and Emek, Alp and Vertens, Johan and Burgard, Wolfram},
  booktitle=icra,
  pages={94--100},
  year={2020}
}

@inproceedings{yuan2025robopoint,
  title={RoboPoint: A Vision-Language Model for Spatial Affordance Prediction in Robotics},
  author={Yuan, Wentao and Duan, Jiafei and Blukis, Valts and Pumacay, Wilbert and Krishna, Ranjay and Murali, Adithyavairavan and Mousavian, Arsalan and Fox, Dieter},
  booktitle=corl,
  pages={4005--4020},
  year={2025},
  organization={PMLR}
}

@techreport{achanta2010slic,
  title={Slic superpixels},
  author={Achanta, Radhakrishna and Shaji, Appu and Smith, Kevin and Lucchi, Aurelien and Fua, Pascal and S{\"u}sstrunk, Sabine and others},
    booktitle={EPFL Technical
Report},
    year={2019}
}

@inproceedings{kirillov2023segment,
  title={Segment anything},
  author={Kirillov, Alexander and Mintun, Eric and Ravi, Nikhila and Mao, Hanzi and Rolland, Chloe and Gustafson, Laura and Xiao, Tete and Whitehead, Spencer and Berg, Alexander C and Lo, Wan-Yen and others},
  booktitle=iccv,
  pages={4015--4026},
  year={2023}
}

@inproceedings{liu2023grounding,
  title={Grounding dino: Marrying dino with grounded pre-training for open-set object detection},
  author={Liu, Shilong and Zeng, Zhaoyang and Ren, Tianhe and Li, Feng and Zhang, Hao and Yang, Jie and Jiang, Qing and Li, Chunyuan and Yang, Jianwei and Su, Hang and others},
  booktitle=eccv,
  pages={38--55},
  year={2024},
  organization={Springer}
}

@article{rampavsek2022recipe,
  title={Recipe for a general, powerful, scalable graph transformer},
  author={Ramp{\'a}{\v{s}}ek, Ladislav and Galkin, Michael and Dwivedi, Vijay Prakash and Luu, Anh Tuan and Wolf, Guy and Beaini, Dominique},
  journal=neurips,
  volume={35},
  pages={14501--14515},
  year={2022}
}

@inproceedings{brohan2023can,
  title={Do as i can, not as i say: Grounding language in robotic affordances},
  author={Brohan, Anthony and Chebotar, Yevgen and Finn, Chelsea and Hausman, Karol and Herzog, Alexander and Ho, Daniel and Ibarz, Julian and Irpan, Alex and Jang, Eric and Julian, Ryan and others},
  booktitle=corl,
  pages={287--318},
  year={2023},
  organization={PMLR}
}

@inproceedings{lai2024lisa,
  title={Lisa: Reasoning segmentation via large language model},
  author={Lai, Xin and Tian, Zhuotao and Chen, Yukang and Li, Yanwei and Yuan, Yuhui and Liu, Shu and Jia, Jiaya},
  booktitle=cvpr,
  pages={9579--9589},
  year={2024}
}

@inproceedings{li2024topviewrs,
  title={TopViewRS: Vision-Language Models as Top-View Spatial Reasoners},
  author={Li, Chengzu and Zhang, Caiqi and Zhou, Han and Collier, Nigel and Korhonen, Anna and Vuli{\'c}, Ivan},
  booktitle=emnlp,
  pages={1786--1807},
  year={2024}
}

@inproceedings{yuksekgonul2023and,
  title={When and why vision-language models behave like bags-of-words, and what to do about it?},
  author={Yuksekgonul, M and Bianchi, F and Kalluri, P and Jurafsky, D and Zou, J and others},
  booktitle=iclr,
  year={2023}
}

@inproceedings{chen2024spatialvlm,
  title={Spatialvlm: Endowing vision-language models with spatial reasoning capabilities},
  author={Chen, Boyuan and Xu, Zhuo and Kirmani, Sean and Ichter, Brain and Sadigh, Dorsa and Guibas, Leonidas and Xia, Fei},
  booktitle=cvpr,
  pages={14455--14465},
  year={2024}
}

@inproceedings{cheng2024spatialrgpt,
  title={SpatialRGPT: grounded spatial reasoning in vision-language models},
  author={Cheng, An-Chieh and Yin, Hongxu and Fu, Yang and Guo, Qiushan and Yang, Ruihan and Kautz, Jan and Wang, Xiaolong and Liu, Sifei},
  booktitle=neurips,
  pages={135062--135093},
  year={2024}
}

@article{elfes2002using,
  title={Using occupancy grids for mobile robot perception and navigation},
  author={Elfes, Albert},
  journal={Computer},
  volume={22},
  number={6},
  pages={46--57},
  year={2002},
  publisher={IEEE}
}

@article{thrun2003learning,
  title={Learning occupancy grid maps with forward sensor models},
  author={Thrun, Sebastian},
  journal={Autonomous robots},
  volume={15},
  number={2},
  pages={111--127},
  year={2003},
  publisher={Springer}
}

@inproceedings{oleynikova2017voxblox,
  title={Voxblox: Incremental 3d euclidean signed distance fields for on-board mav planning},
  author={Oleynikova, Helen and Taylor, Zachary and Fehr, Marius and Siegwart, Roland and Nieto, Juan},
  booktitle=iros,
  pages={1366--1373},
  year={2017},
  organization={IEEE}
}

@inproceedings{oleynikova2016signed,
  title={Signed distance fields: A natural representation for both mapping and planning},
  author={Oleynikova, Helen and Millane, Alexander and Taylor, Zachary and Galceran, Enric and Nieto, Juan and Siegwart, Roland},
  booktitle=rss,
  year={2016},
  organization={University of Michigan}
}

@article{garg2020semantics,
  title={Semantics for robotic mapping, perception and interaction: A survey},
  author={Garg, Sourav and S{\"u}nderhauf, Niko and Dayoub, Feras and Morrison, Douglas and Cosgun, Akansel and Carneiro, Gustavo and Wu, Qi and Chin, Tat-Jun and Reid, Ian and Gould, Stephen and others},
  journal={Foundations and Trends{\textregistered} in Robotics},
  volume={8},
  number={1--2},
  pages={1--224},
  year={2020},
  publisher={Now Publishers, Inc.}
}

@article{grinvald2019volumetric,
  title={Volumetric instance-aware semantic mapping and 3D object discovery},
  author={Grinvald, Margarita and Furrer, Fadri and Novkovic, Tonci and Chung, Jen Jen and Cadena, Cesar and Siegwart, Roland and Nieto, Juan},
  journal=ral,
  volume={4},
  number={3},
  pages={3037--3044},
  year={2019},
  publisher={IEEE}
}

@inproceedings{zhu2023superpixel,
  title={Superpixel transformers for efficient semantic segmentation},
  author={Zhu, Alex Zihao and Mei, Jieru and Qiao, Siyuan and Yan, Hang and Zhu, Yukun and Chen, Liang-Chieh and Kretzschmar, Henrik},
  booktitle=iros,
  pages={7651--7658},
  year={2023},
  organization={IEEE}
}

@article{yang2024rgb,
  title={RGB-D visual odometry by constructing and matching features at superpixel level},
  author={Yang, Meiyi and Xiong, Junlin and Li, Youfu},
  journal={Robotica},
  volume={42},
  number={8},
  pages={2619--2634},
  year={2024},
  publisher={Cambridge University Press}
}

@misc{pybullet,
  author={Coumans, Erwin and Bai, Yunfei},
title = {PyBullet, a Python module for physics simulation for games, robotics and machine learning},
howpublished = {\url{http://pybullet.org}},
year = {2016}
}

\end{document}